# Combining Word Feature Vector Method with the Convolutional Neural Network for Slot Filling in Spoken Language Understanding

Ruixi Lin
CloudMinds Technology Inc.
ruixi.lin@cloudminds.com

**Abstract**

Spoken language understanding (SLU) is an important problem in natural language processing, which involves identifying a user's intent and assigning a semantic concept to each word in a sentence. This paper presents a word feature vector method and combines it into the convolutional neural network (CNN). We consider 18 word features and each word feature is constructed by merging similar word labels. By introducing the concept of external library, we propose a feature set approach that is beneficial for building the relationship between a word from the training dataset and the feature. Computational results are reported using the ATIS dataset and comparisons with traditional CNN as well as bi-directional sequential CNN are also presented.

## 1. Introduction

Spoken language understanding (SLU) is an important problem in natural language processing, which involves identifying a user's intent and assigning a semantic concept to each word in a sentence. For instance, the sentence "I want a cheapest airfare from Tacoma to Orlando" contains two city names, and an SLU system should tag Tacoma as the departure city and Orlando as the arrival city. Many papers have achieved state-of-the-art results in this research area.

Typically, intent detection and slot filling are carried out separately. Logistic regression and support vector machines can be used for intent detection. While traditional approaches for the slot filling task include hidden Markov models (HMM) [1], and conditional random field (CRF) [2]. Recently, neural network models, such as recurrent neural networks (RNNs) and convolutional neural networks (CNNs) have been applied to SLU [3, 4, 5, 6]. A novel neural network model that consists of two RNN was proposed for learning meaningful representations of linguistic phrases [7]. In [8], several extensions of the original RNN language model were presented. CNNs were implemented for natural language processing in [9, 10]. A novel CNN architecture for sequence labeling outperformed even the previously best ensembling recurrent neural network model [11].

Some researchers devote to the jointly training of intent and slot models and have achieved promising results. Triangular CRF (TriCRF) [12] was proposed for this purpose, where an additional random variable indicating the topic of the sentence was introduced. A neural network version of the TriCRF model was established in [13], whose slot filling component was a globally normalized CRF style model and features were automatically extracted through CNN layers.

The remainder of this paper is organized as follows. Section 2 gives a brief introduction to the training dataset and introduces the concepts of feature, external library and feature set. The word's feature vector generation method is described in Section 3. Section 4 reports the computational results. Finally, conclusions are drawn in Section 5.

## 2. Data preprocessing and feature set generation

This section first gives a brief introduction to the training dataset and then presents some preprocessing methods for the dataset. After that, we describe the concept of feature set and external library to facilitate the proposed approach discussed in next section.

(1) Training dataset

We adopt the widely used ATIS dataset [14, 15] as the training dataset. This dataset has a total of 4978 sentences, 56590 words and 572 unduplicated words (including non-word symbols, see Appendix 1). The average length of the sentences is 11.37 words and the number of all labels is 127. According to our analysis, the labels correspond to 1121 key words with a repetition rate of 1121/572=1.96. Names of the labels and their corresponding numbers of key words are listed in Appendix 1.

(2) Feature

Based on the similarity of word labels, we further classify the labels into 18 features (an average of 127/18 = 7 labels in each feature). For example, label "B-city_name" (label index = 17), "B-fromloc.city_name" (label index = 48), "B-stoploc.city_name" (label index = 71) and "B-toloc.city_name" (label index = 78) can be grouped into feature set $S^{city\_name\_1}$. Feature "other" contains the label indexes that are not included in the first 17 features. The feature names with their subordinative label indexes are listed in Table 1.

Table 1 Feature names and their subordinative label indexes.

| No. | Feature Name | Label Indexes | $|S_j|$ | No. | Feature Name | Label Indexes | $|S_j|$ |
|---|---|---|---|---|---|---|---|
| 1 | city_name_1 | 17,48,71,78, | 437 | 10 | airline_code | 1 | 13 |
| 2 | city_name_2 | 91, 109, 119,123 | 95 | 11 | airport_code | 3,46, 69,76 | 9 |
| 3 | state_name_1 | 50,68,81, | 48 | 12 | month_name | 8,28,55,62 | 12 |
| 4 | state_name_2 | 110,118,124 | 11 | 13 | day_number | 7,23,27,61,85,94,115 | 42 |
| 5 | airline_name_1 | 2 | 18 | 14 | day_name | 6,22,26,60 | 14 |
| 6 | airline_name_2 | 83 | 7 | 15 | period_of_day | 12,33,57,65,87,97 | 20 |
| 7 | airport_name_1 | 4,47,70,77 | 121 | 16 | am_pm | 86,88,89,96,98,99,120 | 3 |
| 8 | airport_name_2 | 84, 108, 122 | 87 | 17 | O_set | 126 | 330 |
| 9 | class_type | 18,37 | 6 | 18 | other | remainder | 141 |

(3) External library

We introduce the concept of external library (or dictionary) to help to build the relationship between the words in the dataset and the features. An external library consists of standard words from external resources, for example, a day number library includes seven days of a week, i.e. from Monday to Sunday.

Without loss of generality, we select 474 cities who have airports in the U.S. Among them, 355 cities have single-word city names and the remainder (119 cities) have multi-words city names. Single-word city names and the first words of multi-words city names are classified into feature "city_name_1" (i.e. external library of "city_name_1"), and all remaining words of multi-words city names are grouped into feature "city_name_2". Please note that several cities may share a common word of city names, for instance, city names "san antonio", "san francisco" and "san jose" share the common word of "san". Therefore, the size of a city name feature set is less than the total number of cities it actually contains. The same case can be applied to other feature sets. Similarly, we choose 51 U.S. states and 120 airports to build corresponding external libraries. We also build external libraries for features "class_type", "airline_code" and so on.

(4) Feature set generation

After combining the words from the label sets to the external libraries, we obtain the keyword sets for all the features. In this way, all words from the training dataset can be found in the feature set. One may argue that if a different dataset is used, some words in the dataset cannot be found. To address this issue, the external libraries and the feature sets should keep maintained.

There are a total of 1414 keywords and an average of 78.56 keywords in the feature sets. The fourth and eighth columns of Table 1 present the total number (denoted by $|S_j|$ in Table 1) of keywords for each feature.

## 3. An approach to generating the word's feature vector

We define $a_{jm}$ to describe whether feature group $j$ includes label $m$. Under this definition, the total number of labels that a feature includes can be calculated by the following equation:

$$n_j = \sum_{m=1}^{N^{label}} a_{jm} \quad j=1,2,...,N \quad (1)$$

Furthermore, we define $x_{ij}$ to describe whether word $i$ involves feature $j$ or not. Note that a word could involve more than one features simultaneously. For example, the word "boston" involves both features of $S^{city\_name\_1}$ and $S^{airport\_name\_1}$. We also define $f_i = (\beta_{i1}, \beta_{i2},...,\beta_{iN})$ as the feature vector for word $i$, where $\beta_{ij}$ is the $j$ th dimension of the vector. By introducing the concept of feature weight and normalization, the $j$ th component of the normalized feature vector of word $i$ is calculated as follows:

$$\beta'_{ij} = \frac{\lambda_j n_j x_{ij}}{\sum_{j=1}^{N} \lambda_j n_j x_{ij}} \quad j=1,2,...,N \quad (2)$$

To get a better understanding the computation process of word's feature vector, we use the sample sentence "can you list all flights from washington to toronto" to give more details (see Fig 1). First, we need to determine the relationship between the words and the features. As mentioned above, $x_{ij}=1$ if word $i$ involves feature $j$ and $x_{ij}=0$ otherwise. Second, weights need to be given to all the features. We assume that the features have a homogeneous weigh vector, i.e. all the weights are identical (for example, equal to 1.0).

Next, by substituting the values of $x_{ij}$, $n_j$ and $\lambda_j$ into Eq. (2), we can finally obtain the normalized word's feature vector. For example, the feature vector of word "washington" is $\beta = (0.29, 0.00, 0.21, 0.00, 0.00, 0.00, 0.29, 0.21, 0.00, 0.00, 0.00, 0.00, 0.00, 0.00, 0.00, 0.00)$.

# 4. Computational experiments and discussions

The overall solution framework is depicted in Fig. 2. Indeed, the framework primarily consists of two modules: convolutional neural network training and word's feature vector generation.

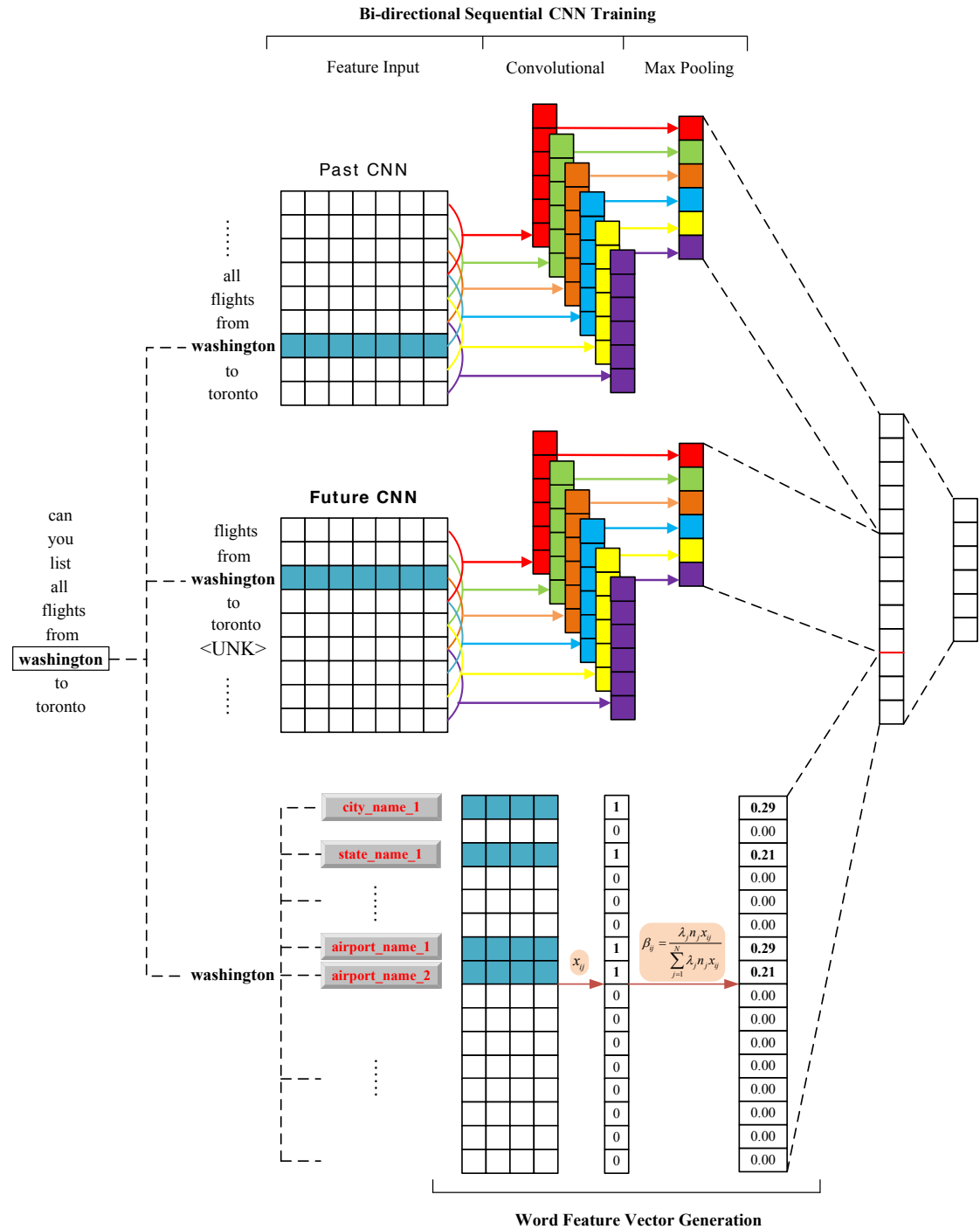

**Fig. 1** The overall solution framework.

The parameter values for the convolutional neural network are set as follows: number of convolutions is 1, activation function is relu, number of filters is 100, size of filter is 100*5, step size of the convolutional process is 1, pooling method is the max pooling operation, dimensions of words are 100 and context length is 7.

Table 2 presents the computational results from three methods: past sequential CNN, bi-directional sequential CNN and past sequential CNN + feature vector (proposed method). We focus on the following four important indicators in terms of the solution quality.

(1) Accuracy (*A*)

$$A = \frac{TP+TN}{TP+TN+FP+FN} \quad (3)$$

(2) Precision (*P*)

$$P = \frac{TP}{TP+FP} \quad (4)$$

(3) Recall (*R*)

$$R = \frac{TP}{TP+FN} \quad (5)$$

(4) F1-score (*F1*)

$$F1 = \frac{2 \times P \times R}{P+R} \quad (6)$$

where TP, TN, FP and FN mean True Positive, True Negative, False Positive and False Negative, respectively.

**Table 4** F1-score of different methods (FYI).

| Method | With/without <UNK> | Embedding size = 100 | Embedding size = 300 |
|---|---|---|---|
| CNN | × | 70.05% | 90.21% |
| | √ | 82.75% | 95.37% |
| CNN + Feature vector (This paper) | × | 83.57% | 91.42% |
| | √ | 92.13% | 95.99% |
| Bi-sCNN | × | 74.08% | 91.44% |
| | √ | 85.44% | 95.75% |
| Bi-sCNN + Feature vector (This paper) | × | 84.34% | 92.16% |
| | √ | 92.35% | 96.36% |

**Table 4** F1-score of different methods.

| Method | Embedding size = 100 | Embedding size = 300 |
|---|---|---|
| CNN | 70.05% | 90.21% |
| CNN + Feature vector (This paper) | 83.57% | 91.42% |
| Bi-sCNN | 74.08% | 91.44% |
| Bi-sCNN + Feature vector (This paper) | 84.34% | 92.16% |

We conduct five experiments with the same dataset and settings and we present the accuracy and F1-score indicators in Table 4. As can be seen from the table, both indicators of the proposed method outperform the past sequential CNN method and bi-directional sequential CNN method. Specifically, in terms of average performance, the proposed method have an improvement of 16.2% on accuracy and 20.7% on F1-score compared with the past sequential CNN method. Meanwhile, an improvement of 5.8% on accuracy and 7.4% on F1-score have achieved compared with the bi-directional sequential CNN method.

# 5. Conclusions

This paper presents a word feature vector method and combines it into the convolutional neural network (CNN). We consider 18 word features and each word feature is constructed by merging similar word labels. By introducing the concept of external library, we propose a feature set approach that is beneficial for building the relationship between a word from the training dataset and the feature. A series computational experiments are carried out to validate our method. Computational results indicate that our proposed method outperforms two traditional methods (namely the past sequential CNN and bi-directional sequential CNN) in terms of the indicators of accuracy and F1-score.